# Translating Federated Learning Algorithms in Python into CSP Processes Using ChatGPT

Miroslav Popovic *, Marko Popovic **, Miodrag Djukic *, Ilija Basicevic *
* University of Novi Sad, Faculty of Technical Sciences, Novi Sad, Serbia
** RT-RK Institute for Computer Based Systems, Novi Sad, Serbia
E-mail address of the corresponding author: miroslav.popovic@rt-rk.uns.ac.rs

***Abstract* — The Python Testbed for Federated Learning Algorithms is a simple Python FL framework that is easy to use by ML&AI developers who do not need to be professional programmers and is also amenable to LLMs. In the previous research, generic federated learning algorithms provided by this framework were manually translated into the CSP processes and algorithms' safety and liveness properties were automatically verified by the model checker PAT. In this paper, a simple translation process is introduced wherein the ChatGPT is used to automate the translation of the mentioned federated learning algorithms in Python into the corresponding CSP processes. Within the process, the minimality of the used context is estimated based on the feedback from ChatGPT. The proposed translation process was experimentally validated by successful translation (verified by the model checker PAT) of both generic centralized and decentralized federated learning algorithms.**

## I. Introduction

This research was conducted within the ongoing EU Horizon 2020 project TaRDIS [1] that aims to create a toolbox for easy programming of AI/ML powered distributed swarm applications, which is to be validated on the four industrial use cases: (1) smart grids, (2) smart telcos, (3) smart robotized factories, and (4) smart LEO satellite constellations. Besides being smart, these systems are safety-critical, and therefore they need to be correct-by-construction i.e., they need to be formally verified.

Python Testbed for Federated Learning Algorithms (PTB-FLA) [2] is a simple FL framework targeting edge systems that provides both generic centralized and decentralized FL algorithms, which implement the corresponding FL orchestration protocols. To aid easy programming in tune with low-code/no-code initiative, PTB-FLA offers: (1) a simple API that is amenable both to nonprofessional developers and LLMs such as ChatGPT [3], (2) the 4-phase development paradigm for humans [4] and (3) the adapted 4-phase and 2-phase development paradigms for ChatGPT [5].

Recently, the PTB-FLA FL orchestration protocols were formally verified in the 2-phases process by using the process algebra Communicating Sequential Processes (CSP) and the model checker PAT [6]. The main limitation of [6] is that Python implementations of PTB-FLA FL orchestration protocols were manually translated into the corresponding CSP processes using an ad hoc approach. Motivated by the goal to overcome this limitation, authors of [7] devised a process in which they: (1) first rewrite the original Python code using the restricted actor-based programming model, and then (2) systematically construct CSP code from the rewritten Python code.

This paper was motivated by the idea to see whether it would be possible to skip rewriting the Python code, and to directly translate the original Python code into the corresponding CSP processes using contemporary LLMs. We decided to use ChatGPT (concretely ChatGPT-4o) as we already had a very positive experience with it and as it seems that ChatGPT is generally the most popular AI tool with its 100 million users as of Nov. 19th, 2024 [8].

Recent works in ML society have demonstrated that GPT-3.5 and GPT-4 (or GPT for short) have capabilities in planning and coding, which was not exhibited in the previous ML models [9], [10], [11]. There are many examples of using these GPT-4 capabilities. For example, authors of [12] propose autonomous AI edge framework, wherein the requests from users are expressed in natural language, which is initially processes by GPT to understand and route them to the appropriate edge AI models. The goal of this paper is to translate Python code into CSP processes, so it mostly uses GPT coding capabilities.

In this paper, a simple translation process is introduced wherein the ChatGPT is used to automate the translation of the federated learning algorithms in Python into the corresponding CSP processes. The proposed translation process was experimentally validated by successful translation (verified by the model checker PAT) of both PTB-FLA generic centralized and decentralized federated learning algorithms.

The main paper contributions are the following: (1) the automated translation process for translating PTB-FLA generic federated learning algorithms in Python into CSP processes, (2) the input contexts (prompts) given to ChatGPT, (3) the CSP models of both PTB-FLA generic centralized and decentralized federated learning algorithms (FLAs) produced by this translation process, and (4) the feedback questionnaires completed by ChatGPT that are used to estimate the minimality of the used contexts.

Funded by the European Union (TaRDIS, 101093006). Views and opinions expressed are however those of the author(s) only and do not necessarily reflect those of the European Union. Neither the European Union nor the granting authority can be held responsible for them.

The rest of the paper is organized as follows. Section II presents the translation process, Section III presents the experimental validation, and Section IV concludes the paper.

## II. TRANSLATION PROCESS

In this paper, the translation process is defined as an automated process of transforming the given PTB-FLA federated learning algorithm in Python into the corresponding CSP processes by leveraging ChatGPT. The output CSP processes are written in a dialect of CSP called CSP#, which is supported by the model checker PAT [13]. Further in the text, the terms output CSP processes and the CSP# code are used interchangeably.

The translation process is a simple iterative process, which comprises the following steps: (1) specify the context (i.e., the prompt), (2) ask ChatGPT to complete the assignment, (3) test the output CSP# code by the model checker PAT, (4) if the CSP# code successfully passes testing, stop, else if the CSP# code is close to what is expected and contains small number of simple errors, manually modify the code and return to step 3, else return to step 1. To successfully pass testing by PAT, the output CSP# code must pass two steps: (3.1) it must be syntactically correct and (3.2) all the specified (desired) system properties must be successfully verified (PAT uses the term validated instead).

We tried to create minimal contexts to minimize the overall translation effort/cost (because the ChatGPT service is charged by the context size). For the sake of this research, to get an estimation of the context quality (minimality), after the translation process was successfully completed, we asked ChatGPT for feedback by asking the following three questions: (i) On the scale from 1 to 10, where 10 is the hardest, how would you rate this assignment? (ii) What was the most important part of the context that you used to complete the assignment? and (iii) What was the redundant part of the context?

## III. EXPERIMENTAL VALIDATION

This section presents the experimental validation of the translation process for the two cases: (1) the translation of the PTB-FLA generic centralized FLA and (2) the translation of the PTB-FLA generic decentralized FLA. For each case two tables are given: (1) the context and (2) the CSP# code (generated by ChatGPT that was manually tweaked to make it pass testing by PAT) with the ChatGPT feedback. The text in italic in Table I is reused in Table III (see the comments in italic and enclosed by angle brackets in Table III).

The next subsections present the experimental setup, the case of the centralized FLA, and the case of the decentralized FLA, respectively.

### A. Experimental setup

ChatGPT has three control parameters: *top-k*, *temperature*, and *top-p*. The *top-k* controls how many top-ranked tokens are considered for each word prediction (a lower value increases determinism, while a higher value can enhance creativity). The *temperature* adjusts the randomness of token selection (a value of 1 is neutral, higher values increase randomness, and lower values make the output more focused). The *top-p* (nucleus sampling) controls the cumulative probability mass, only considering tokens whose combined probability reaches this threshold.

The experimental validation was conducted with the ChatGPT default setting, which is as follows: (1) *top-k* = 40, (2) *temperature* = 0.7, and (3) *top-p* = 0.9. As declared by ChatGPT, these values aim to make a balance between generating varied, creative responses and staying coherent and contextually relevant.

### B. Translation of PTB-FLA Generic Centralized FLA

Table I consist of the three parts: (1) problem statement, MPAPI description, and MPAPI translation to CSP#, (2) the skeleton of the target CSP# model comprising definitions of constants, variables, channels, system, and system properties, and (3) the input Python code of the function fl_centralized to be translated to CSP#.

TABLE I. CONTEXT FOR CENTRALIZED FLA

/*

*We are considering a distributed algorithms written as Python functions, and we are interested in translating these functions into the corresponding #CSP models. (CSP# is a dialect of CSP that was developed specifically for the Process Analysis Toolkit (PAT), a model checker created at the National University of Singapore.)*

*Each of these Python functions is executed by noNodes nodes (or processes). Each node has its own identification nodeId. If the algorithm is centralized, flSrvId is the server identification. Nodes communicate by messaging – each node (or actor) receives messages from its local queue and sends messages to other nodes' remote queues. In CSP#, we may model these queues as an array of channels, which is indexed by the nodeId.*

*A typical argument of the Python function is some input data. In CSP#, we may model data for all the nodes as a shared memory array, which is then indexed by the nodeId. There may be more input data arguments, and consequently more var arrays in the #CSP model. The output data i.e., the return values may be modelled analogously.*

*Following the restricted programming approach, the Python function performs messaging by using the given message-passing API (MPAPI) comprising the following four functions: sendMsg(address, msg), rcvMsg(), broadcastMsg(addresses, msg, senderId), and rcvMsgs(noMsgs).*

The translation of sendMsg and rcvMsg calls to #CSP depends on the format of messages.

If the message has a single field, than the call sendMsg (address, msg) translates to nodeChannels [nodeId]!msg, and the call msg=rcvMsg() translates to nodeChannels[nodeId]?msg. If the message has more fields, say 3 fields, then the call sendMsg (address, [m1, m2, m3]) translates to nodeChannels [nodeId]!m1.m2.m3, and the call [m1, m2, m3]=rcvMsg() translates to nodeChannels [nodeId]?m1.m2.m3.

*The function broadcastMsg uses the function sendMsg, and the function rcvMsgs uses the function rcvMsg. Both broadcastMsg and rcvMsgs perform loops which may be transformed into the tail recursion. The translation of broadcastMsg and rcvMsgs calls depends on the message format in the same way as translation of sendMsg and rcvMsg calls.*

For example, the call broadcastMsg(addresses, msg, senderId), where the msg is a single field message, translates to:

```
broadcastMsg(msg, senderId) =
  broadcastMsgT(0, NoNodes, msg, senderId);
broadcastMsgT(i, noNodes, msg, senderId) =
```

```
  if (i < noNodes) {
    if (i != senderId) {
         nodeChannels[i]!msg -> Skip
    };
    broadcastMsgT(i+1, noNodes, msg, senderId)
  };
```

Analogously, the call msgs=rcvMsgs(noMsgs), which receives noMsgs and stores them in the Python list msgs, translates to (in CSP#, the list msgs is modeled as the shared array msgs):

```
channel buffer[NoNodes] NoNodes-1;
rcvMsgs(nodeId, noMsgs) =
  if (bufferSize[nodeId] != 0) {
    clearBuffT(nodeId)
  };
  rcvMsgsT(0, nodeId, noMsgs);
clearBuffT(nodeId) =
  {bufferSize[nodeId]--} ->
  buffer[nodeId]?temp ->
  if (bufferSize[nodeId] != 0) {
    clearBuffT(nodeId)
  };
rcvMsgsT(i, nodeId, noMsgs) =
  if (i < noMsgs) {
    {bufferSize[nodeId]++} ->
    nodeChannels[nodeId]?temp ->
    buffer[nodeId]!temp ->
    rcvMsgsT(i+1, nodeId, noMsgs)
  };
```

*After this general introduction made above, we focus on a particular Python function fl_centralized(nodeId, ldata, pdata), where ldata and pdata are its local and private data, respectively. The system comprising NoNode instances of this function may be modeled in CSP# as follows:*

```
*/
// The constants, variables, and channels
enum {False, True};
#define NoNodes 3;
#define FlSrvId 0;
#define NoIterations 3;
var ldataArr[NoNodes];
var pdataArr[NoNodes];
var terminated = False;
channel nodeChannels[NoNodes] NoNodes-1;
// The system
SysCentralized() = |||nodeId:{0..NoNodes-1}
  @Fl_centralized(nodeId, ldataArr[nodeId], pdataArr[nodeId]);
// Properties
#assert SysCentralized() deadlockfree;
#define Terminated (terminated == True);
#assert SysCentralized() reaches Terminated;
#assert SysCentralized() |= []<> Terminated;
/*
The input Python function:
def fl_centralized(nodeId, localData, privateData):
  for k in range(noIterations):
    if nodeId == flSrvId:
      # Server
      broadcastMsg(addresses, localData, nodeId)
      msgs = rcvMsgs(noNodes-1)
    else:
      # Client
      msg = rcvMsg()
      sendMsg(flSrvAddress, localData)
  terminated = 1
*/
Fl_centralized(nodeId, ldata, pdata) =
```

The translation was finished in the first iteration. Table II shows the final CSP# code and the ChatGPT feedback. ChatGPT made the same syntax error by using the terminal “->” instead of “;” in two lines, which were corrected by replacing “;” for “->” (see the comments in Tabel II). In the feedback, ChatGPT: (1) scored difficulty as 7 out of 10, (2) found the MPAPI explanation as the most crucial, and (3) as expected found the multi-filed message example (not used in fl_centralized) as redundant.

TABLE II. RESULT FOR CENTRALIZED FLA

```
Fl_centralized(nodeId, ldata, pdata) =
  Fl_centralizedT(0, nodeId, ldata, pdata);
Fl_centralizedT(k, nodeId, ldata, pdata) =
  if (k < NoIterations) {
    // Server behavior
    if (nodeId == FlSrvId) {
      broadcastMsg(ldata, nodeId);  // syntax error "->" instead of ";"
      rcvMsgs(nodeId, NoNodes - 1); // syntax error "->" instead of ";"
      Fl_centralizedT(k+1, nodeId, ldata, pdata)
    }
    // Client behavior
    else {
      nodeChannels[nodeId]?msg ->
      nodeChannels[FlSrvId]!ldata ->
      Fl_centralizedT(k+1, nodeId, ldata, pdata)
    }
  } else {
    {terminated = True} ->
    Skip
  };
```

Here are the answers to your questions:

1. Rating: I would rate this assignment 7 out of 10. < *Shortened.* >

2. Most Important Part of the Context: The most crucial part of the context was the message-passing API (MPAPI) explanation, which provided the translation rules for sendMsg, rcvMsg, broadcastMsg, and rcvMsgs. This API allowed me to map the communication patterns in the Python function into CSP#.

3. Redundant Part of the Context: The multi-field message example wasn't directly necessary for this specific translation. < *Shortened.* >

### C. *Translation of PTB-FLA Generic Decentralized FLA*

Table III consists of the same three parts as Table I. Some of the text is reused or analogous to italic text parts in Table I and the new text comprises: (1) the new definitions of constants, variables, and channels, and (2) the input Python code of the function fl_decentralized to be translated to CSP#.

TABLE III. CONTEXT FOR DECENTRALIZED FLA

/*

< *The first 4 paragraphs are the same as in Table I.* >

The message msg is encoded in Python as a list whose elements correspond to message fields. The list msg translates to the compound message in CSP# which is constructed by concatenating adjacent message fields with dots, e.g., the message '[m1, m2, m3, m4]' in Python gets translated into the compound message 'm1.m2.m3.m4' in CSP#.

The call 'sendMsg(address, msg)' in Python translates to 'nodeChannels[nodeId]!msg' in CSP#. For example, the call 'sendMsg(address, [m1, m2, m3, m4])' translates to 'nodeChannels[nodeId]!m1.m2.m3.m4' in CSP#.

The call 'msg = rcvMsg()' in Python translates to 'nodeChannels[nodeId]?msg' in CSP#. For example, if the message is a list with four elements named m1, m2, m3, and m4, the call 'msg = rcvMsg()' translates to 'nodeChannels[nodeId]?m1.m2.m3.m4'.

< Here goes the copy of the 7th paragraph form Table I. >

For example, if the message msg is a list with four elements named m1, m2, m3, and m4, the call 'broadcastMsg(addresses, msg, senderId)', translates to the following CSP# code (note that in the translated code, the argument 'addresses' maps to the indices of the array of channels nodeChannels):

```
BroadcastMsg(m1, m2, m3, m4, senderId) =
  BroadcastMsgT(0, NoNodes, m1, m2, m3, m4, senderId);
BroadcastMsgT(i, noNodes, m1, m2, m3, m4, senderId) =
  if (i < noNodes) {
    if (i != senderId) {
      nodeChannels[i]! m1.m2.m3.m4 -> Skip
  };
  BroadcastMsgT(i + 1, noNodes, m1, m2, m3, m4, senderId)
};
```

Besides using the API described above, Python functions may have their own temporary FIFO buffers implemented as lists and use list operations to append an element at the list tail or to pop an element from the list head. In CSP#, such lists translate to channels, and operations on lists translate to sending and receiving messages over channels. For example, if the Python list named buf is used as an FIFO buffer, and if the buf is a local variable for the node with the identification nodeId, and if the message msg is a list with four elements named m1, m2, m3, and m4, then the Python expression 'buf.append(msg)' translates into the CSP# expression 'buf[nodeId]!m1.m2.m3.m4, whereas the Python expression 'msg = buf.pop(0)' translates into the CSP# expression 'buf[nodeId]?m1.m2.m3.m4.

< *Here go the lines analogous to the lines in italic in Table I.* >

*/

// The constants, variables, and channels

< *Here go the same lines in italic from Table I.* >

```
#define NoNeighbors NoNodes-1;
#define PHASE1 1;
#define PHASE2 2;
var noRcvdMsgs[NoNodes]; // inits to 0
channel nodeChannels[NoNodes] 2*(NoNodes-1);
channel dataFromClients1[NoNodes] NoNodes-1;
channel dataFromClients2[NoNodes] NoNodes-1;
```

// The system

< *Here go the lines analogous to the lines in italic in Table I.* >

/*

The input python function:

```
def fl_decentralized(nodeId, localData, privateData):
  #   The nodes exchange messages, which are lists with 4 elements.
  #   The indices of these message elements are the following:
  msgIterNo = 0
  msgSeqNo = 1
  msgSrcAdr = 2
  msgData = 3
  #   The msgSeqNo message field can have two values:
  PHASE1 = 1
  PHASE2 = 2
  # Note: missing variables and lists are defined elsewhere
  for iterNo in range(noIterations):
    # This node is initially acting as a server - phase 1
    broadcastMsg(addresses,
      [iterNo, PHASE1, nodeId, localData], nodeId)
    # This node now acts as a client - phase 2
    noRcvdMsgs = 0
    # First drain the buffer dataFromClients1
    while len(dataFromClients1) > 0:
      msg = dataFromClients1.pop(0)
      noRcvdMsgs = noRcvdMsgs + 1
      sendMsg(msg[msgSrcAdr], [iterNo, PHASE2, nodeId, localData])
    # Process the rest of the messages
    while noRcvdMsgs != 2*noNeighbors:
      msg = rcvMsg()
      # Message from a client already in the iteration iterNo+1?
      if msg[msgIterNo] != iterNo:
        dataFromClients1.append(msg)
        continue
      noRcvdMsgs = noRcvdMsgs + 1
      if msg[msgSeqNo] == PHASE1:
        # The 1st msg from a neighbor acting as a server
```

```
      # This node takes the role of a client
      sendMsg(msg[msgSrcAdr],
        [iterNo, PHASE2, nodeId, localData])
    else:
      # The 2nd msg from a neighbor acting as client
      # This node takes the role of a server
      dataFromClients2.append(msg[msgData])
  # All 2*noNeighbors messages have been processed
  # This node takes the final role of a server - phase 3
  # Consume all the msgs from the dataFromClients2 - drop them
  dropMsgsFromClients2(dataFromClients2)
# All the iterations are completed
terminated = 1
*/
Fl_decentralized(nodeId, ldata, pdata) =
```

The translation was finished in the fourth iteration. Table IV shows the final CSP# code and the ChatGPT feedback. Overall, ChatGPT made four syntax errors and two logical (i.e., semantic) errors (see the comments in Tabel IV). As in Table II, ChatGPT made the same syntax error by using the terminal "->" instead of ";" in two lines (see comments for the syntax errors 1 and 3 in Table IV), which were corrected by replacing ";" for "->". Then ChatGPT didn't put curly brackets around the statement "terminated = True" (the syntax error 2 in Table IV). Finally, ChatGPT didn't put the parameter ldata in the definition of the process DrainBuffer (the syntax error 4 in Table IV). The corrections of the last two syntax errors are explained in Table IV.

Besides the syntax errors, ChatGPT made the two logical errors: (1) the process DrainBuffer was wrongly called from Fl_decentralized_Phase2 and it should have been called from Fl_decentralizedT and (2) the line "noRcvdMsgs[nodeId]++} ->" was missing in DrainBuffer. Unlike the four syntax errors that were easy to correct, it took a significant effort to uncover the two logical errors as the only means to do it was analysis of PAT log files or PAT simulation traces (and both seem rather cryptic to users not too familiar with PAT).

In the feedback, ChatGPT: (1) scored difficulty as 8 out of 10, (2) found the explanation of how to handle specific message-passing functions and variables as the most crucial, and (3) found the initial description of the broadcastMsg and rcvMsgs functions as largely redundant and that explanations about standard CSP# syntax were less necessary after the first few translations (but this remark is to be taken with our objection that without these explanations the CSP# code in the first translation iteration was very far from the expected one).

TABLE IV. RESULT FOR DECENTRALIZED FLA

```
Fl_decentralized(nodeId, ldata, pdata) =
  Fl_decentralizedT(nodeId, ldata, pdata, 0); // Start at iteration 0
// Recursive process to handle each iteration
Fl_decentralizedT(nodeId, ldata, pdata, iterNo) =
  if (iterNo < NoIterations) {
    // Phase 1: Server broadcasts to clients
    // Syntax error 1: In the line below "->" instead of ";"
    BroadcastMsg(iterNo, PHASE1, nodeId, ldata, nodeId);
    // Reset noRcvdMsgs for this node at the start of each iteration
    { noRcvdMsgs[nodeId] = 0 } ->
    // Phase 2: Acts as client and processes incoming messages
    /* Logical error 1 correction: the following call was moved from the
beginning of Fl_decentralized_Phase2 here. */
    DrainBuffer(nodeId, iterNo, ldata,
      call(ccount, dataFromClients1[nodeId]));
    Fl_decentralized_Phase2(nodeId, ldata, pdata, iterNo)
  } else {
    // Syntax error 2: In the line below was: "terminated = True;"
    {terminated = True} -> // After all iterations, mark termination
    Skip
  };
  // Phase 2: Processes msgs until 2*noNeighbors msgs received
Fl_decentralized_Phase2(nodeId, ldata, pdata, iterNo) =
/* Syntax error 3: In the line below "->" instead of ";".
Syntax error 4: In the same line, parameter ldata was missing, Correction
1: ldata was added before the parameter count.
Logical error 1: Draining dataFromClients1 should be done only once per
iteration. Correction was to move this call to the calling process. */
  /*DrainBuffer(nodeId, iterNo, ldata, call(ccount,
dataFromClients1[nodeId])); */
  // Do until noRcvdMsgs[nodeId] reaches 2 * NoNeighbors
  if (noRcvdMsgs[nodeId] < 2 * NoNeighbors) {
    nodeChannels[nodeId]?msgIterNo.msgSeqNo.msgSrcAdr.msgData ->
    if (msgIterNo != iterNo) {
      // Store message for the next iteration if it's from the next iteration
      dataFromClients1[nodeId]!
        msgIterNo.msgSeqNo.msgSrcAdr.msgData ->
      Fl_decentralized_Phase2(nodeId, ldata, pdata, iterNo)
    } else {
      // Increment the received messages count for this node
      { noRcvdMsgs[nodeId] = noRcvdMsgs[nodeId] + 1 } ->
      if (msgSeqNo == PHASE1) {
        // Respond to phase 1 messages
        nodeChannels[msgSrcAdr]!iterNo.PHASE2.nodeId.ldata ->
        Fl_decentralized_Phase2(nodeId, ldata, pdata, iterNo)
      } else {
        // Store data for phase 2 messages
        dataFromClients2[nodeId]!msgData ->
        Fl_decentralized_Phase2(nodeId, ldata, pdata, iterNo)
      }
    }
  } else {
```

```
  // Final role of server, phase 3
  DropMsgsFromClients2(nodeId, call(ccount,
   dataFromClients2[nodeId]));
  Fl_decentralizedT(nodeId, ldata, pdata, iterNo + 1)
 };
 /* Drain the buffer dataFromClients1 by forwarding each message to nodeChannels[msgSrcAdr]. */
 /* Syntax error 4, Correction 2: parameter ldata was missing, so it was added before the parameter count. */
DrainBuffer(nodeId, iterNo, ldata, count) =
 if (count > 0) {
  dataFromClients1[nodeId]?
   msgIterNo.msgSeqNo.msgSrcAdr.msgData ->
  // Logical error 2: The line below was missing
   noRcvdMsgs[nodeId]++} ->
  // Syntax error 4: In line below ldata was used without declaration.
  nodeChannels[msgSrcAdr]!iterNo.PHASE2.nodeId.ldata ->
  /* Syntax error 4, Correction 3: parameter ldata was missing, so it was added before the parameter count. */
  DrainBuffer(nodeId, iterNo, ldata, count - 1)
 } else {
  Skip
 };
```

Thank you for your feedback! Here's my response to your questions:

1. Difficulty Rating: I'd rate this assignment an 8. Translating procedural Python logic into CSP# requires a detailed understanding of CSP# syntax, its limitations, and the different ways to handle iteration, recursion, and state updates. Managing this within CSP#'s limited syntax without typical Python constructs (like direct variable updates or list handling) added complexity.

2. Most Important Context: The explanation of how to handle specific message-passing functions and variables (like noRcvdMsgs, dataFromClients1, and dataFromClients2) was key. These instructions shaped how to translate the Python code, especially in terms of converting Python's list operations into CSP# message channels and recursive processes.

3. Redundant Context: The initial description of the broadcastMsg and rcvMsgs functions was largely redundant. Once their structure was understood, the repetitive parts of those descriptions weren't as critical for later translation tasks. Similarly, explanations about standard CSP# syntax were less necessary after the first few translations.

## IV. Conclusion

In this paper, ChatGPT is used in a simple translation process to automate translating PTB-FLA generic federated learning algorithms (FLAs) in Python into the corresponding CSP processes. The translation process was experimentally validated by successful translation of both PTB-FLA generic centralized and decentralized FLAs.

The main advantage of this paper over the previous papers [6], [7] is that part of the work is done by ChatGPT. The second advantage of this paper over [7] is that it takes the original Python code as its input, whereas [7] requires the Python code to be rewritten in the restricted actor-based programming model.

The main limitation of this paper is that humans must manually correct all the errors made by ChatGPT. The main direction of the future work is to overcome this limitation. One approach to do it is to create a dataset for this specific translation process to reduce the number of errors. Another approach is to use an open LLM and a tool like IterGen [14] to eliminate the syntax errors and perhaps be able to at least partly cope with the logical errors.